\def\year{2020}\relax
\documentclass[letterpaper]{article} % DO NOT CHANGE THIS
\usepackage{aaai20} % DO NOT CHANGE THIS
\usepackage{times} % DO NOT CHANGE THIS
\usepackage{helvet} % DO NOT CHANGE THIS
\usepackage{courier} % DO NOT CHANGE THIS
\usepackage[hyphens]{url} % DO NOT CHANGE THIS
\usepackage{graphicx} % DO NOT CHANGE THIS
\usepackage{graphicx} % DO NOT CHANGE THIS
\usepackage{url}
\usepackage{graphicx} %Required
\usepackage{amsmath}
\usepackage{amsfonts}

\newcommand{\argmax}{\operatornamewithlimits{argmax}}

\newcommand{\softmax}{\operatornamewithlimits{softmax}}
\title{Conclusion-Supplement Answer Generation for Non-Factoid Questions}
\author{Makoto Nakatsuji, Sohei Okui\\
NTT Resonant Inc.\\
Granparktower, 3-4-1 Shibaura,
Minato-ku, Tokyo 108-0023, Japan
\\
nakatsuji.makoto@gmail.com, okui@nttr.co.jp
}

\begin{document}

\maketitle

\begin{abstract}
This paper tackles the goal of conclusion-supplement answer generation
for non-factoid questions, which is a critical issue in the field of
Natural Language Processing (NLP) and Artificial Intelligence (AI), as
users often require supplementary information before accepting a
conclusion. The current encoder-decoder framework, however, has
difficulty generating such answers, since it may become confused when it
tries to learn several different long answers to the same non-factoid
question. Our solution, called an {\em ensemble network}, goes beyond
single short sentences and fuses logically connected conclusion
statements and supplementary statements.
%; the key novelty is that it
%influences the supplement decoder with the conclusion decoder output via
%attention mechanism and influences both the conclusion and the
%supplement decoder with the question encoder as: (1) 
It extracts {\em the context} from the conclusion decoder's output
sequence and uses it to create supplementary decoder states on the basis
of an attention mechanism.
%(2) 
It also assesses {\em the closeness} of the question encoder's output
sequence and the separate outputs of the conclusion and supplement
decoders as well as {\em their combination}. As a result, it generates
answers that match the questions and have natural-sounding supplementary
sequences in line with the context expressed by the conclusion
sequence. 
Evaluations conducted on datasets including ``Love Advice'' and ``Arts \& Humanities'' categories indicate that our model outputs much more accurate results than the tested baseline models do.
\end{abstract}

\section{Introduction}
Question Answering (QA) modules play particularly important roles in
recent dialog-based Natural Language Understanding (NLU) systems, such
as Apple's Siri and Amazon's Echo. Users chat with AI systems in natural
language to get the answers they are seeking. QA systems can deal with
two types of question: factoid and non-factoid ones. The former sort
asks, for instance, for the name of a thing or person such as ``What/Who
is $X$?''. The latter sort includes more diverse questions that cannot
be answered by a short fact. For instance, users may ask for advice on
how to make a long-distance relationship work well or for opinions on
public issues. Significant progress has been made in answering factoid
questions
\cite{wang-smith-mitamura:2007:EMNLP-CoNLL2007,DBLP:journals/corr/YuHBP14};
however, answering non-factoid questions remains a challenge for QA
modules.

Long short term memory (LSTM) sequence-to-sequence models
\cite{Sutskever:2014:SSL:2969033.2969173,DBLP:journals/corr/VinyalsL15,BahdanauCB14}
try to generate short replies to the short utterances often seen in chat
systems. Evaluations have indicated that these models have the
possibility of supporting simple forms of general knowledge QA,
e.g. ``Is the sky blue or black?'', since they learn commonly occurring
sentences in the training corpus. Recent machine reading comprehension
(MRC) methods
\cite{DBLP:conf/nips/NguyenRSGTMD16,DBLP:journals/corr/RajpurkarZLL16}
try to return a single short answer to a question by extracting answer
spans from the provided passages. Unfortunately, they may generate
unsatisfying answers to regular non-factoid questions because they can
easily become confused when learning several different long answers to
the same non-factoid question, as pointed out by
\cite{jia-liang-2017-adversarial,wang-etal-2018-multi-passage}.

This paper tackles a new problem: conclusion-supplement answer
generation for non-factoid questions. Here, the conclusion consists of
sentences that directly answer the question, while the supplement
consists of information supporting the conclusion, e.g., reasons or
examples. Such conclusion-supplement answers are important for helping
questioners decide their actions, especially in NLU. As described in
\cite{ennis}, users prefer a supporting supplement before accepting an
instruction (i.e., a conclusion). Good debates also include claims
(i.e., conclusions) about a topic and supplements to support them that
will allow users to reach decisions
\cite{DBLP:conf/emnlp/RinottDPKAS15}. The following example helps to
explain how conclusion-supplement answers are useful to users: ``Does
separation by a long distance ruin love?'' Current methods tend to
answer this question with short and generic replies, such as, ``Distance
cannot ruin true love''. The questioner, however, is not likely to be
satisfied with such a trite answer and will want to know how the
conclusion was reached. If a supplemental statement like ``separations
certainly test your love'' is presented with the conclusion, the
questioner is more likely to accept the answer and use it to reach a
decision. Furthermore, there may be multiple answers to a non-factoid
question. For example, the following answer is also a potential answer
to the question: ``distance ruins most relationships. You should keep in
contact with him''. The current methods, however, have difficulty
generating such conclusion-supplement answers because they can become
easily confused when they try to learn several different and long
answers to a non-factoid question.

To address the above problem, we propose a novel architecture, called
the {\em ensemble network}. It is an extension of existing
encoder-decoder models, and it generates two types of decoder output
sequence, conclusion and supplement. It uses two viewpoints for
selecting the conclusion statements and supplementary
statements. (Viewpoint 1) {\em The context} present in the conclusion
decoder's output is linked to supplementary-decoder output states on the
basis of an attention mechanism. Thus, the context of the conclusion
sequence directly impacts the decoder states of the supplement
sequences. This, as a result, generates natural-sounding supplementary
sequences. (Viewpoint 2) {\em The closeness} of the question sequence
and conclusion (or supplement) sequence as well as the closeness of the
question sequence with {\em the combination} of conclusion and
supplement sequences is considered. By assessing the closeness at the
sentence level and sentence-combination level in addition to at the word
level, it can generate answers that include good supplementary sentences
following the context of the conclusion. This avoids having to learn
several different conclusion-supplement answers assigned to a single
non-factoid question and generating answers whose conclusions and
supplements are logically inconsistent with each other.

Community-based QA (CQA) websites tend to provide answers composed of
conclusion and supplementary statements; from our investigation, 77\% of
non-factoid answers (love advice) in the Oshiete-goo
(https://oshiete.goo.ne.jp) dataset consist of these two statement
types. The same is true for 82\% of the answers in the Yahoo non-factoid
dataset\footnote{https://ciir.cs.umass.edu/downloads/nfL6/} related to
the fields of social science, society \& culture and arts \&
humanities.
%\footnote{We randomly sampled 1,000 answers from each dataset
%and checked if the answers consisted of conclusions and/or
%supplements. If the answer contained only information that directly
%answered the question, it was of the conclusion type. If there was
%information that supported the conclusion, e.g. reasoning, the answer
%was of the conclusion with supplement type.}. 
We used the
above-mentioned CQA datasets in our evaluations, since they provide
diverse answers given by many responders. The results showed that our
method outperforms existing ones at generating correct and natural
answers. We also conducted an love advice
service\footnote{http://oshiete.goo.ne.jp/ai} in Oshiete goo to evaluate
the usefulness of our ensemble network.

\section{Related work}
The encoder-decoder framework learns how to transform one representation
into another. Contextual LSTM (CLSTM) incorporates contextual features
(e.g., topics) into the encoder-decoder framework
\cite{DBLP:journals/corr/GhoshVSRDH16,DBLP:conf/aaai/SerbanSBCP16}. It
can be used to make the context of the question a part of the answer
generation process. HieRarchical Encoder Decoder (HRED)
\cite{DBLP:conf/aaai/SerbanSBCP16} extends the hierarchical recurrent
encoder-decoder neural network into the dialogue domain; each question
can be encoded into a dense context vector, which is used to recurrently
decode the tokens in the answer sentences. Such sequential generation of
next statement tokens, however, weakens the original meaning of the
first statement (question). Recently, several models based on the
Transformer \cite{vaswani2017attention}, such as for passage ranking
\cite{DBLP:journals/corr/abs-1904-08375,DBLP:journals/corr/abs-1804-07888}
and answer selection \cite{article}, have been proposed to evaluate
question-answering systems. There are, however, few Transformer-based
methods that generate non-factoid answers.

\begin{figure*}[t]
\begin{center}
\includegraphics[width=\linewidth]{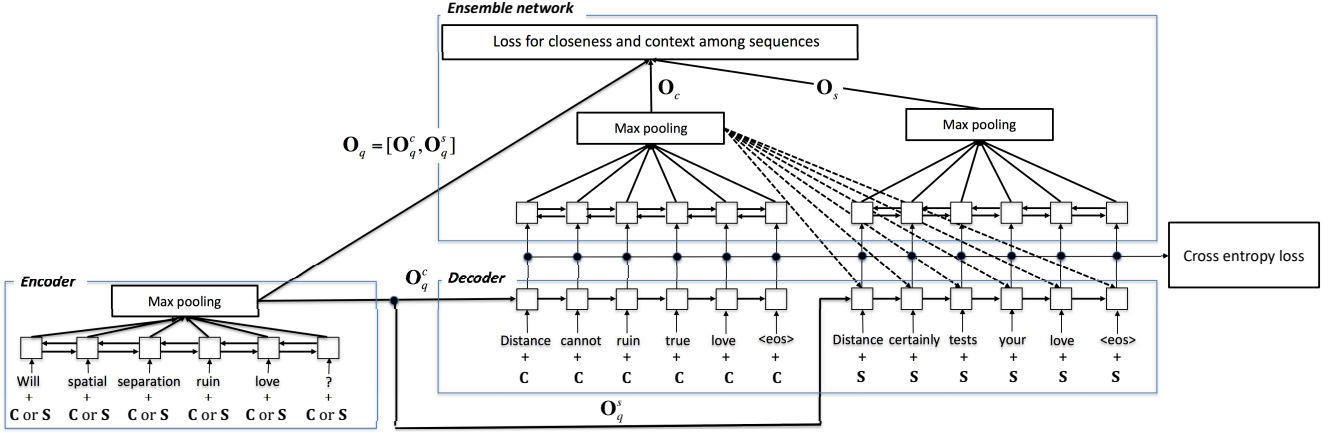}
\end{center}
\caption{Neural conclusion-supplement answer generation model. } 
\label{fig:1}
\vspace{-2mm}
\end{figure*}

Recent neural answer selection methods for non-factoid questions
\cite{dossantos-EtAl:2015:ACL-IJCNLP,Qiu,TanSXZ16} learn question and
answer representations and then match them using certain similarity
metrics. They use open datasets stored at CQA sites like Yahoo! Answers
since they include many diverse answers given by many responders and
thus are good sources of non-factoid QA training data. The above
methods, however, can only select and extract answer sentences, they do
not generate them.

Recent machine reading comprehension methods try to answer a question
with exact text spans taken from provided passages
\cite{wei2018fast,DBLP:journals/corr/RajpurkarZLL16,DBLP:conf/emnlp/YangYM15,DBLP:journals/corr/JoshiCWZ17}. Several
studies on the MS-MARCO dataset
\cite{DBLP:journals/corr/TanWYLZ17,DBLP:conf/nips/NguyenRSGTMD16,wang-etal-2018-multi-passage}
define the task as using multiple passages to answer a question where
the words in the answer are not necessarily present in the
passages. Their models, however, require passages other than QA pairs
for both training and testing. Thus, they cannot be applied to CQA
datasets that do not have such passages. Furthermore, most of the
questions in their datasets only have a single answer. Thus, we think
their purpose is different from ours; generating answers for non-factoid
questions that tend to demand diverse answers.

There are several 
%QA methods for the non-factoid questions raised in
complex QA tasks such as those present in the TREC complex interactive
QA tasks\footnote{https://cs.uwaterloo.ca/~jimmylin/ciqa/} or
DUC\footnote{http://www-nlpir.nist.gov/projects/duc/guidelines.html}
complex QA tasks. Our method can be applied to those non-factoid
datasets if an access fee is paid.
%\cite{TanSXZ16,conf/ecir/YangASCPCGS16,weko_293_1,Chaturvedi:2014:JQC:2566486.2567999}

\section{Model}
This section describes our conclusion-supplement answer generation model
in detail. An overview of its architecture is shown in Figure
\ref{fig:1}.

Given an input question sequence ${\bf{Q}} = \{{\bf{q}}_1, \cdots ,
{\bf{q}}_i, \cdots , {\bf{q}}_{N_q}\}$, the proposal outputs a
conclusion sequence ${\bf{C}} = \{{\bf{c}}_1, \cdots, {\bf{c}}_t, \cdots
, {\bf{c}}_{N_c}\}$, and supplement sequence ${\bf{S}} = \{{\bf{s}}_1,
\cdots, {\bf{s}}_t, \cdots , {\bf{s}}_{N_s}\}$. The goal is to learn a
function mapping from ${\bf{Q}}$ to ${\bf{C}}$ and ${\bf{S}}$. Here,
${\bf{q}}_i$ denotes a one-of-$K$ embedding of the $i$-th word in an
input sequence of length $N_q$. ${\bf{c}}_t$ (${\bf{s}}_t$) denotes a
one-of-$K$ embedding of the $t$-th word in an input sequence of length
$N_c$ ($N_s$).

\subsection{Encoder}
The encoder converts the input $\bf{Q}$ into a question embedding,
${\bf{O}}_q$, and hidden states, ${\bf{H}}={\{{\bf{h}}_i\}_i}$.

Since the question includes several pieces of background information on
the question, e.g. on the users' situation, as well as the question
itself, it can be very long and composed of many sentences. For this
reason, we use the BiLSTM encoder, which encodes the question in both
directions, to better capture the overall meaning of the question. It
processes both directions of the input, $\{{\bf{q}}_1, \cdots ,
{\bf{q}}_{N_q}\}$ and $\{{\bf{q}}_{N_q}, \cdots , {\bf{q}}_{1}\}$,
sequentially. At time step $t$, the encoder updates the hidden state by:
\begin{equation*}
\begin{split}
 {\bf{h}}_i &= [{\bf{h}}^f_i, {\bf{h}}^b_i]^{\mathrm{T}}
 \mbox{ s.t. }\\
 {\bf{h}}^f_i &= f({\bf{q}}_{i-1}, {\bf{h}}^f_{i-1}), {\bf{h}}^b_i
 = f({\bf{q}}_{i+1}, {\bf{h}}^b_{i+1}),
\end{split} \label{eq:ht}
\end{equation*}
where $f()$ is an LSTM unit, and ${\bf{h}}^f_i$ and ${\bf{h}}^b_i$ are
hidden states output by the forward-direction LSTM and
backward-direction LSTM, respectively.

We also want to reflect sentence-type information such as conclusion
type or supplement type in sequence-to-sequence learning to better
understand the conclusion or supplement sequences. We achieve this by
adding a sentence type vector for conclusion $\bf{C}$ or for supplement
$\bf{S}$ to the input gate, forget gate output gate, and cell memory
state in the LSTM model. This is equivalent to processing a composite
input [${\bf{q}}_i$, $\bf{C}$] or [${\bf{q}}_i$, $\bf{S}$] in the LSTM
cell that concatenates the word embedding and sentence-type embedding
vectors. We use this modified LSTM in the above BiLSTM model as:

\begin{equation*}
\begin{split}
{\bf{h}}_i &= [{\bf{h}}^f_i, {\bf{h}}^b_i]^{\mathrm{T}} \mbox{ s.t. }\\
 {\bf{h}}^f_i &= f([{\bf{q}}_{i-1}, {\bf{C}}], {\bf{h}}^f_{i-1}), 
 {\bf{h}}^b_i
 = f([{\bf{q}}_{i+1}, {\bf{C}}], {\bf{h}}^b_{i+1}).
\end{split}
\end{equation*}
When encoding the question to decode the supplement sequence, ${\bf{S}}$ is input instead of ${\bf{C}}$ in the above equation.

The BiLSTM encoder then applies a max-pooling layer to all hidden vectors to extract the most salient signal for each word. As a result, it generates a fixed-sized distributed vector representation for the conclusion, ${\bf{O}}^c_q$, and another for the supplement, ${\bf{O}}^s_q$. ${\bf{O}}^c_q$ and ${\bf{O}}^s_q$ are different since the encoder is biased by the corresponding sentence-type vector, $\bf{C}$ or $\bf{S}$.
 
As depicted in Figure \ref{fig:1}, the BiLSTM encoder processes each word with a sentence-type vector (i.e. $\bf{C}$ or $\bf{S}$) and the max-pooling layer to produce the question embedding ${\bf{O}}^c_q$ or ${\bf{O}}^s_q$. These embeddings are used as context vectors in the decoder network for the conclusion and supplement.

\subsection{Decoder}
The decoder is composed of a conclusion decoder and supplement decoder. Here, let ${\bf{h}}^{\prime}_t$ be the hidden state of the $t$-th LSTM unit in the conclusion decoder. Similar to the encoder, the decoder also decodes a composite input [${\bf{c}}_t$, $\bf{C}$] in an LSTM cell that concatenates the conclusion word embedding and sentence-type embedding vectors. It is formulated as follows:

\begin{equation*}
\begin{split}
{\bf{h}}^{\prime}_t &= f^{\prime}([{\bf{c}}_{t-1}, {\bf{C}}],
 {\bf{h}}^{\prime}_{t-1}) \mbox{ s.t. } \\ {{\bf{c}}_{t-1}} &= \argmax_{c}
 \softmax_c({\bf{h}}^{\prime}_{t-1}),
\end{split}
\end{equation*}
where $f^{\prime}()$ denotes the conclusion decoder LSTM, $\softmax_c$ the probability of word $c$ given by a softmax layer, $c_t$ the $t$-th conclusion decoded token, and ${\bf{c}}_t$ the word embedding of $c_t$. The supplement decoder's hidden state ${\bf{h}}^{\prime\prime}_t$ is computed in the same way with ${\bf{h}}^{\prime}_t$; however, it is updated in the ensemble network described in the next subsection.

As depicted in Figure \ref{fig:1}, the LSTM decoder processes tokens according to question embedding ${\bf{O}}^c_q$ or ${\bf{O}}^s_q$, which yields a bias corresponding to the sentence-type vector, $\bf{C}$ or $\bf{S}$. The output states are then input to the ensemble network.

\subsection{Ensemble network}
The conventional encoder-decoder framework often generates short and simple sentences that fail to adequately answer non-factoid questions. Even if we force it to generate longer answers, the decoder output sequences become incoherent when read from the beginning to the end.

The ensemble network solves the above problem by (1) passing {\em the context} from the conclusion decoder's output sequence to the supplementary decoder hidden states via an attention mechanism, and (2) considering {\em the closeness} of the encoder's input sequence to the decoders' output sequences as well as the closeness of the encoder's input sequence to {\em the combination} of decoded output sequences.

(1) To control the {\em context}, we assess all the information output by the conclusion decoder and compute the conclusion vector, ${\bf{O}}_c$. ${\bf{O}}_c$ is a sentence-level representation that is more compact, abstractive, and global than the original decoder output sequence. To get it, we apply BiLSTM  to the conclusion decoder's output states $\{ {{{\tilde{\bf{y}}}}_t^c} \}_t$; i.e., $\{ {{{\tilde{\bf{y}}}}_t^c} \}_t = \{{\bf{U}}\cdot \softmax({\bf{h}}^{\prime}_t)\}_t$, where word representation matrix $\bf{U}$ holds the word representations in its columns. At time step $t$, the BiLSTM encoder updates the hidden state by:
\begin{equation*}
 \begin{split}
{\bf{h}}^c_t \;\; &= [{\bf{h}}^{c,f}_t, {\bf{h}}^{c,b}_t]^{\mathrm{T}}
 \mbox{ s.t. } \\
 {\bf{h}}^{c,f}_t &= f({{{\tilde{\bf{y}}}}_{t-1}^c}, {\bf{h}}^{c,f}_{t-1}), \;\;\; {\bf{h}}^{c,b}_t
 = f({{{\tilde{\bf{y}}}}_{t+1}^c}, {\bf{h}}^{c,b}_{t+1}),
\end{split}
\end{equation*}
where ${\bf{h}}^{c,f}_t$ and ${\bf{h}}^{c,b}_t$ are the hidden states output by the forward LSTM and backward LSTM in the conclusion encoder, respectively. It applies a max-pooling layer to all hidden vectors to extract the most salient signal for each word to compute the embedding for conclusion ${\bf{O}}_c$. Next, it computes the context vector ${\bf{cx}}_t$ at the $t$-th step by using the $(t\!\!-\!\!1)$-th output hidden state of the supplement decoder, ${\bf{h}}^{\prime\prime}_{t\!-\!1}$, weight matrices, ${\bf{V}}_a$ and ${\bf{W}}_a$, and a sigmoid function, $\sigma$:
\begin{equation*}
{\bf{cx}}_t = \alpha_{t} {{\bf{O}}_c} \; \mbox{ s.t. }\;
\alpha_{t} = \sigma({\bf{V}}_a^{\mathrm{T}} \tanh({\bf{W}}_a {\bf{h}}^{\prime\prime}_{t\!-\!1} + {\bf{O}}_c)).
\end{equation*}

This computation lets our ensemble network extract a conclusion-sentence level context. The resulting supplement sequences follow the context of the conclusion sequence. Finally, ${{\bf{h}}}^{\prime\prime}_t$ is computed as:

\begin{small}
\begin{eqnarray}\label{eq:ot}
{\bf{z}}_t &=& \sigma({\bf{W}}_z [{\bf{y}}_{t-1}, {\bf{T}}] + {\bf{U}}_z {\bf{h}}^{\prime\prime}_{t-1}
 + {{\bf{W}}^a_z} {{\bf{cx}}_t} + {\bf{b}}_z)\\ 
\widetilde{{\bf{l}}}_t &=& \tanh({\bf{W}}_l [{\bf{y}}_{t-1}, {\bf{T}}] + {\bf{U}}_l {\bf{h}}^{\prime\prime}_{t-1} + {{\bf{W}}^a_l} {{\bf{cx}}_t}+
 {\bf{b}}_l)  \nonumber \\
{{\bf{l}}}_t &=& {\bf{i}}_t \ast \widetilde{{\bf{l}}}_t + {\bf{f}}_t \ast {{\bf{l}}_{t-1}}  \nonumber\\ \nonumber
{\bf{h}}^{\prime\prime}_t &=& {\bf{o}}_t \ast \tanh({\bf{l}}_t)
\end{eqnarray}
\end{small}
$z$ can be $i$, $f$, or $o$, which represent three gates (e.g., input
${\bf{i}}_t$, forget ${\bf{f}}_t$, and output
${\bf{o}}_t$). ${\bf{l}}_t$ denotes a cell memory
vector. ${{\bf{W}}}^a_z$ and ${{\bf{W}}}^a_l$ denote attention
parameters.

(2) To control the {\em closeness} at the sentence level and {\em sentence-combination} level, it assesses all the information output by the supplement decoder and computes the supplement vector, ${\bf{O}}_s$, in the same way as it computes ${\bf{O}}_c$. That is, it applies BiLSTM to the supplement decoder's output states $\{ {{{\tilde{\bf{y}}}}_t^s} \}_t$; i.e., $\{ {{{\tilde{\bf{y}}}}_t^s} \}_t = \{{\bf{U}}\!\cdot\! \softmax({{\bf{h}}_t^{\prime\prime}})\}_t$, where the word representations are found in the columns of $\bf{U}$. Next, it applies a max-pooling layer to all hidden vectors in order to compute the embeddings for supplement ${\bf{O}}_s$. Finally, to generate the conclusion-supplement answers, it assesses {\em the closeness} of the embeddings for the question ${\bf{O}}_q$ to those for the answer sentences (${\bf{O}}_c$ or ${\bf{O}}_s$) and {\em their combination}  ${\bf{O}}_c$ and ${\bf{O}}_s$. The loss function for the above metrics is described in the next subsection.

As depicted in Figure \ref{fig:1}, the ensemble network computes the conclusion embedding ${\bf{O}}_c$, the attention parameter weights from ${\bf{O}}_c$ to the decoder output supplement states (dotted lines represent attention operations), and the supplement embedding ${\bf{O}}_s$. Then, ${\bf{O}}_c$ and ${\bf{O}}_s$ are input to the loss function together with the question embedding ${\bf{O}}_q = [{\bf{O}}^c_q,{\bf{O}}^s_q]$.

\subsection{Loss function of ensemble network}
Our model uses a new loss function rather than generative supervision, which aims to maximize the conditional probability of generating the sequential output $p({\bf{y}}|{\bf{q}})$. This is because we think that assessing the closeness of the question and an answer sequence as well as the closeness of the question to two answer sequences is useful for generating natural-sounding answers.

The loss function is for optimizing the closeness of the question and conclusion and that of the question and supplement as well as for optimizing the closeness of the question with the combination of the conclusion and supplement. The training loss ${\cal{L}}_s$ is expressed as the following hinge loss, where ${\bf{O}}^{+}$ is the output decoder vector for the ground-truth answer, ${\bf{O}}^{-}$ is that for an incorrect answer randomly chosen from the entire answer space, $M$ is a constant margin, and $\mathbb{A}$ is set equal to $\{[{\bf{O}}^{+}_c, {\bf{O}}^{-}_s], [{\bf{O}}^{-}_c, {\bf{O}}^{+}_s], [{\bf{O}}^{-}_c, {\bf{O}}^{-}_s]\}$:

\begin{small}
\begin{eqnarray*}
 {\cal{L}}_s =
\sum_{{\bf{O}}_a \in \mathbb{A}}{ \max\{0,
 M\!\!-\!\!(\cos({\bf{O}}_q, [{\bf{O}}^{+}_c, {\bf{O}}^{+}_s]) \!\!-\!
 \cos({\bf{O}}_q , {\bf{O}}_a))\} } \nonumber
\end{eqnarray*}
\end{small}

The key idea is that ${\cal{L}}_s$ checks whether or not the conclusion, supplement, and their combination have been well predicted. In so doing, ${\cal{L}}_s$ can optimize not only the prediction of the conclusion or supplement but also the prediction of the combination of conclusion and supplement.

The model is illustrated in the upper part of Figure \ref{fig:1}; $({\bf{O}}_q, {\bf{O}}_c, {\bf{O}}_s)$ is input to compute the closeness and sequence combination losses.

\subsection{Training}
The training loss ${\cal{L}}_w$ is used to check ${\cal{L}}_s$ and the cross-entropy loss in the encoder-decoder model.  In the following equation, the conclusion and supplement sequences are merged into one sequence $\bf{Y}$ of length $T$, where $T\!=\!N_c\!+\!N_s$. 

\begin{eqnarray}
 {\cal{L}}_w = \alpha \cdot {\cal{L}}_s - \ln{\prod_{t=1}^T
 p({\bf{y}}_t|\bf{Q},{\bf{y}}_1,\ldots,{\bf{y}}_{t-1})}. \label{eq:alpha}
\end{eqnarray}
$\alpha$ is a parameter to control the weighting of the two losses. We use adaptive stochastic gradient descent (AdaGrad) to train the model in an end-to-end manner. The loss of a training batch is averaged over all instances in the batch.
 
Figure \ref{fig:1} illustrates the loss for the ensemble network and the cross-entropy loss.

\section{Evaluation}

\subsection{Compared methods}
\label{sec:comparedMethods} 

We compared the performance of our method with those of (1) {\em
Seq2seq}, a seq2seq attention model proposed by \cite{BahdanauCB14}; (2)
{\em CLSTM}, i.e., the CLSTM model
\cite{DBLP:journals/corr/GhoshVSRDH16}; (3) {\em Trans}, the Transformer
\cite{vaswani2017attention}, which has proven effective for common NLP
tasks. In these three methods, conclusion sequences and supplement
sequences are decoded separately and then joined to generate
answers. They give more accurate results than methods in which the
conclusion sequences and supplement sequences are decoded
sequentially. We also compared (4) {\em HRED}, a hierarchical recurrent
encoder-decoder model \cite{DBLP:conf/aaai/SerbanSBCP16} in which
conclusion sequences and supplement sequences are decoded sequentially
to learn the context from conclusion to supplement; (5) {\em NAGMWA},
i.e., our neural answer generation model without an attention
mechanism. This means that {\em NAGMWA} does not pass ${\bf{cx}}_t$ in
Eq. (\ref{eq:ot}) to the decoder, and conclusion decoder and supplement
decoder are connected only via the loss function ${\cal{L}}_s$. In the
tables and figures that follow, {\em NAGM} means our full model.

\subsection{Dataset}
Our evaluations used the following two CQA datasets:

\paragraph{Oshiete-goo}
The Oshiete-goo dataset includes questions stored in the ``love
advice'' category of the Japanese QA site, Oshiete-goo. It has 771,956
answers to 189,511 questions. We fine-tuned the model using a corpus
containing about 10,032 question-conclusion-supplement (q-c-s)
triples. We used 2,824 questions from the Oshiete-goo dataset. On
average, the answers to these questions consisted of about 3.5
conclusions and supplements selected by human experts. The questions,
conclusions, and supplements had average lengths of 482, 41, and 46
characters, respectively. There were 9,779 word tokens in the questions
and 6,317 tokens in answers; the overlap was 4,096.

\paragraph{nfL6}
We also used the Yahoo nfL6 dataset, the largest publicly available
English non-factoid CQA dataset. It has 499,078 answers to 87,361
questions. We fine-tuned the model by using questions in the ``social
science'', ``society \& culture'', and ``arts \& humanities''
categories, since they require diverse answers. This yielded 114,955
answers to 13,579 questions. We removed answers that included some stop
words, e.g. slang words, or those that only refer to some URLs or
descriptions in literature, since such answers often become noise when
an answer is generated. Human experts annotated 10,299
conclusion-supplement sentences pairs in the answers.

In addition, we used a neural answer-sentence classifier to classify the
sentences into conclusion or supplement classes. It first classified the
sentences into supplements if they started with phrases such as ``this
is because'' or ``therefore''. Then, it applied a BiLSTM with
max-pooling to the remaining unclassified sentences, ${\bf{A}} =
\{{\bf{a}}_1, {\bf{a}}_2, \cdots , {\bf{a}}_{N_a}\}$, and generated
embeddings for the un-annotated sentences, ${\bf{O}}^a$. After that, it
used a logistic sigmoid function to return the probabilities of mappings
to two discrete classes: conclusion and supplement. This mapping was
learned by minimizing the classification errors using the above 10,299
labeled sentences. As a result, we automatically acquired 70,000
question-conclusion-supplement triples from the entire answers. There
were 11,768 questions and 70,000 answers. Thus, about 6 conclusions and
supplements on average were assigned to a single question. The
questions, conclusions, and supplements had average lengths of 46, 87,
and 71 characters, respectively.
%
%\footnote{Japanese tend to write very
%long questions, unlike Americans. We think this reflects a cultural
%difference.}.  
We checked the performance of the classifier; human
experts checked whether the annotation results were correct or not. They
judged that it was about 81\% accurate (it classified 56,762 of 70,000
sentences into correct classes). There were 15,690 word tokens in
questions and 124,099 tokens in answers; the overlap was 11,353.

{\tabcolsep=1.7mm
\begin{table}[t]
\begin{center}
\caption{Results when changing $\alpha$.}
 \footnotesize
{\tabcolsep = 2mm

 \begin{tabular}{c|c|c|c|c|c|c} 
& \multicolumn{3}{c|}{{\it Oshiete-goo}} & \multicolumn{3}{c}{{\em nfL6}} \\ \hline
$\alpha$ & 0 & 1 & 2 & 0 & 
 1 & 2 \\ \hline\cline{1-3}\cline{1-2}\cline{2-4}\cline{5-7}
ROUGE-L & 0.251 & {\bf{\em 0.299}} & 0.211 & 0.330 & {\bf{\em 0.402}} & 0.295 \\\hline
BLEU-4 & 0.098 & {\bf{\em 0.158}} & 0.074 &0.062  & {\bf{\em 0.181}} & 0.023  \\\hline
 \end{tabular}

}
\label{tab:alphabet}
\end{center}
\vspace{-2mm}
\end{table}
}
 \normalsize

{\tabcolsep=1.7mm
\begin{table}[t]
\begin{center}
\caption{Results when using sentence-type embeddings.}
 \footnotesize
{\tabcolsep = 2mm

 \begin{tabular}{c|c|c|c|c} 
& \multicolumn{2}{c|}{{\it Oshiete-goo}} & \multicolumn{2}{c}{{\em nfL6}} \\ \hline 
 & {\em NAGM} & {\em w/o ste}   & {\em NAGM} & {\em w/o ste}
 \\ \hline\cline{1-2}\cline{3-4}
ROUGE-L &   {\bf{\em 0.299}} & 0.235  & {\bf{\em 0.402}} & 0.349 \\\hline
BLEU-4 &   {\bf{\em 0.158}} & 0.090 & {\bf{\em 0.181}} & 0.067 \\\hline
 \end{tabular}

}
\label{tab:ste}
\end{center}
\vspace{-2mm}
\end{table}
}
 \normalsize

{\tabcolsep=1.7mm
\begin{table}[t]
\begin{center}
\caption{ROUGE-L/BLEU-4 for Oshiete-goo.}
 \footnotesize
{\tabcolsep = 1.0mm
\begin{tabular}{c|ccccccc}
 &{\it Seq2seq} & {\it CLSTM} & {\it Trans} & 
 {\it HRED}& {\it NAGMWA} & {\it NAGM} \\ \hline
ROUGE-L & 0.238 & 0.260 & 0.278 & 0.210 & 0.291 & {\bf{\em 0.299}} \\\hline
BLEU-4 & 0.092 & 0.121 & 0.087 & 0.042 & 0.147 & {\bf{\em 0.158}} \\\hline
\end{tabular}
}
\label{tab:result5}
\end{center}
\vspace{-2mm}
\end{table}
}
 \normalsize

{\tabcolsep=1.7mm
\begin{table}[t]
\begin{center}
\caption{ROUGE-L/BLEU-4 for nfL6.}
 \footnotesize
{\tabcolsep = 1.0mm
\begin{tabular}{c|cccccc}
 &{\it Seq2seq} & {\it CLSTM} & {\it Trans} & 
 {\it HRED}& {\it NAGMWA} & {\it NAGM} \\ \hline
ROUGE-L & 0.291 & 0.374 & 0.338 & 0.180 & 0.383 & {\bf{\em 0.402}} \\\hline
BLEU-4 & 0.081 & 0.141 & 0.122 & 0.055 & 0.157 & {\bf{\em 0.181}} \\\hline
\end{tabular}
}
\label{tab:result555}
\end{center}
\vspace{-2mm}
\end{table}
}
 \normalsize

\begin{table}[t]
 \begin{minipage}[t]{.45\textwidth}
 \begin{center}
 \caption{Human evaluation (Oshiete-goo).}
 \begin{footnotesize}
 \begin{tabular}{c|c|c|c|c|c|c|c} 
 \hline
\multicolumn{4}{c|}{{\it CLSTM}} & \multicolumn{4}{c}{{\em NAGM}} \\ \hline
 (1) & (2) & (3) & (4) & 
 (1) & (2) & (3) & (4)\\ \hline
 21 & 18 & 27 & 34 & {\bf{\em 47}} & 32 & 11 & 10 \\ \hline	 
 \end{tabular}
	\label{tab:result6}
	\end{footnotesize}
 \end{center}
 \end{minipage}
 \hfill
 \begin{minipage}[t]{.45\textwidth}
 \begin{center}
 \caption{Human evaluation (nfL6).}
\begin{footnotesize}
 \begin{tabular}{c|c|c|c|c|c|c|c}
 \hline
\multicolumn{4}{c|}{{\it CLSTM}} & \multicolumn{4}{c}{{\em NAGM}} \\ \hline
(1) & (2) & (3) & (4) & 
 (1) & (2) & (3) & (4)\\ \hline
 30 & 3 & 27 & 40 & {\bf{\em 50}} & 23 & 16 & 11 \\ \hline
 \end{tabular}
 \label{tab:result8}
 \end{footnotesize}
 \end{center}
 \end{minipage}
\vspace{-2mm}
\end{table}
 \normalsize

\subsection{Methodology}
\label{sec:methodology} 

We conducted three evaluations using the Oshiete-goo dataset; we
selected three different sets of 500 human-annotated test pairs from the
full dataset. In each set, we trained the model by using training pairs
and input questions in test pairs to the model. 
%Each evaluation used
%9,532 human-annotated and 500 human-annotated test triples.
We repeated the experiments three times by randomly shuffling the
train/test sets.

For the evaluations using the nfL6 dataset, we prepared three different
sets of 500 human-annotated test q-c-s triples from the full dataset. We
used 10,299 human-annotated triples to train the neural sentence-type
classifier. Then, we applied the classifier to the unlabeled answer
sentences. Finally, we evaluated the answer generation performance by
using three sets of machine-annotated 69,500 triples and 500
human-annotated test triples.

After training, we input the questions in the test triples to the model
to generate answers for both datasets. We compared the generated answers
with the correct answers. The results described below are average values
of the results of three evaluations.

The softmax computation was slow since there were so many word tokens in
both datasets. Many studies
\cite{DBLP:conf/ijcai/YinJLSLL16,DBLP:conf/nips/YangYWCS16,DBLP:journals/corr/VinyalsL15}
restricted the word vocabulary to one based on frequency. This, however,
narrows the diversity of the generated answers. Since diverse answers
are necessary to properly reply to non-factoid questions, we used bigram
tokens instead of word tokens to speed up the computation without
restricting the vocabulary. Accordingly, we put 4,087 bigram tokens in
the Oshiete-goo dataset and 11,629 tokens in the nfL6 dataset.

To measure performance, we used human judgment as well as two popular
metrics
\cite{Sutskever:2014:SSL:2969033.2969173,DBLP:conf/nips/YangYWCS16,BahdanauCB14}
for measuring the fluency of computer-generated text: ROUGE-L
\cite{lin:2004:ACLsummarization} and BLEU-4
\cite{Papineni:2002:BMA:1073083.1073135}. ROUGE-L is used for measuring
the performance for evaluating non-factoid QAs
\cite{Song:2017:SAN:3018661.3018704}, however, we also think human
judgement is important in this task.

\subsection{Parameter setup}
\label{sec:parameter} For both datasets, we tried different parameter
values and set the size of the bigram token embedding to 500, the size
of LSTM output vectors for the BiLSTMs to $500 \times 2$, and number of
topics in the CLSTM model to 15. We tried different margins, $M$, in the
hinge loss function and settled on $0.2$. The iteration count $N$ was
set to $100$.

We varied $\alpha$ in Eq. (\ref{eq:alpha}) from 0 to 2.0 and checked the
impact of $L_s$ by changing $\alpha$. Table \ref{tab:alphabet} shows the
results. When $\alpha$ is zero, the results are almost as poor as those
of the seq2seq model. On the other hand, while raising the value of
$\alpha$ places greater emphasis on our ensemble network, it also
degrades the grammaticality of the generated results. We set $\alpha$ to
1.0 after determining that it yielded the best performance. This result
clearly indicates that our ensemble network contributes to the accuracy
of the generated answers.

A comparison of our full method {\em NAGM} with the one without the
sentence-type embedding (we call this method {\em w/o ste}) that trains
separate decoders for two types of sentences is shown in Table
\ref{tab:ste}. The result indicated that the existence of the sentence
type vector, $\bf{C}$ or $\bf{S}$, contributes the accuracy of the
results since it distinguishes between sentence types.

\subsection{Results}
\label{sec:results}
\paragraph{Performance}
The results for Oshiete-goo are shown in Table \ref{tab:result5} and
those for nfL6 are shown in Table \ref{tab:result555}. They show that
{\em CLSTM} is better than {\em Seq2seq}. This is because it
incorporates contextual features, i.e. topics, and thus can generate
answers that track the question's context. {\em Trans} is also better
than {\em Seq2seq}, since it uses attention from the question to the
conclusion or supplement more effectively than {\em Seq2seq}. {\em HRED}
failed to attain a reasonable level of performance. These results
indicate that sequential generation has difficulty generating subsequent
statements that follow the original meaning of the first statement
(question).

{\em NAGMWA} is much better than the other methods except {\em NAGM},
since it generates answers whose conclusions and supplements as well as
their combinations closely match the questions. Thus, conclusions and
supplements in the answers are consistent with each other and avoid
confusion made by several different conclusion-supplement answers
assigned to a single non-factoid questions. Finally, {\em NAGM} is
consistently superior to the conventional attentive encoder-decoders
regardless of the metric. Its ROUGE-L and BLEU-4 scores are much higher
than those of {\em CLSTM}. Thus, {\em NAGM} generates more fluent
sentences by assessing the context from conclusion to supplement
sentences in addition to the closeness of the question and sentences as
well as that of the question and sentence combinations.

%Despite outperforming the other methods, our method had lower ROUGE-L
%and BLEU-4 scores than those of systems intended for providing simple
%short answers, since non-factoid answers are usually long (e.g. the
%MS-MARCO natural language generation task achieved higher ROUGE-L and
%BLEU-4 scores in their evaluation than we achieved for the nfL6 dataset
%since the nfL6 dataset has 2.3 times longer answers in terms of
%character length than the MS-MARCO dataset has.). The scores, however,
%do correlate with the human evaluation results, as shown in the next
%subsection.

\renewcommand{\figurename}{Table}
\setcounter{figure}{6}

\begin{figure*}[t]
\begin{center}
\caption{Example answers generated by {\it CLSTM} and {\em NAGM}. \#1 is for Oshiete-goo and \#2 for nfL6.}
\includegraphics[width=\linewidth]{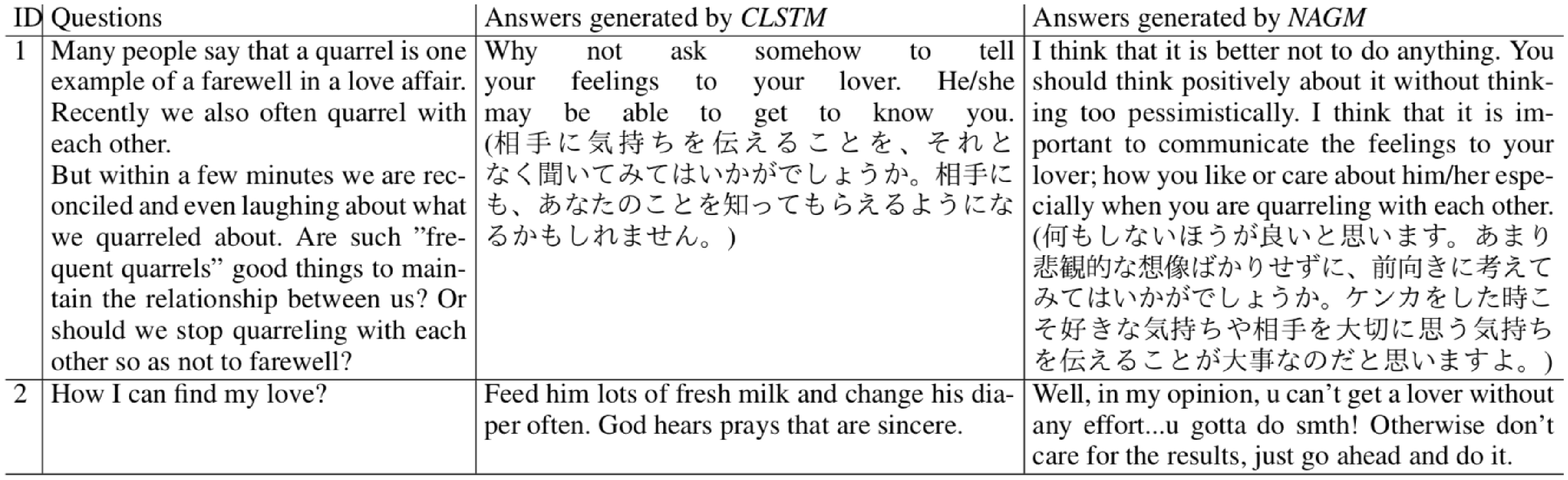}
\end{center}
\label{fig:result7}
\vspace{-2mm}
\end{figure*}

\paragraph{Human evaluation}
Following evaluations made by crowdsourced evaluators \cite{DBLP:conf/emnlp/LiMRJGG16}, we conducted human evaluations to judge the outputs of {\it CLSTM} and those of {\em NAGM}. Different from \cite{DBLP:conf/emnlp/LiMRJGG16}, we hired human experts who had experience in Oshiete-goo QA community service. Thus, they were familiar with the sorts of answers provided by and to the QA community.

The experts asked questions, which were not included in our training datasets, to the AI system and rated the answers; one answer per question. The experts rated the answers as follows: (1) the content of the answer matched the question, and the grammar was okay; (2) the content was suitable, but the grammar was poor; (3) the content was not suitable, but the grammar was okay; (4) both the content and grammar were poor. Note that our evaluation followed the DUC-style strategy\footnote{http://www-nlpir.nist.gov/projects/duc/duc2007/quality-questions.txt}. Here, we mean ``grammar'' to cover grammaticality, non-redundancy, and referential clarity in the DUC strategy, whereas we mean the ``content matched the questions'' to refer to ``focus'' and ``structure and coherence'' in the DUC strategy. The evaluators were given more than a week to carefully evaluate the generated answers, so we consider that their judgments are reliable. Each expert evaluated 50 questions. We combined the scores of the experts by summing them. They did not know the identity of the system in the evaluation and reached their decisions independently.

Table \ref{tab:result6} and Table \ref{tab:result8} present the results. The numbers are percentages. Table 7 presents examples of questions and answers. For Oshiete-goo results, the original Japanese and translated English are presented. The questions are very long and include long background descriptions before the questions themselves. 

These results indicate that the experts were much more satisfied with the outputs of {\em NAGM} than those of {\it CLSTM}. This is because, as can be seen in Table 7, {\it NAGM} generated longer and better question-related sentences than {\it CLSTM} did. {\it NAGM} generated grammatically good answers whose conclusion and supplement statements are well matched with the question and the supplement statement naturally follows the conclusion statement.
 
\paragraph{Generating answers missing from the corpus}
The encoder-decoder network tends to re-generate answers in the training corpus. On the other hand, {\em NAGM} can generate answers not present in the corpus by virtue of its ensemble network that considers contexts and sentence combinations.

Table 7 lists some examples. For example, answer \#1 generated by {\em NAGM} is not in the training corpus. We think it was generated from the parts in italics in the following three sentences that are in the corpus: (1) ``{\em I think that it is better not to do anything} from your side. If there is no reaction from him, it is better not to do anything even if there is opportunity to meet him next.'' (2) ``I think it may be good for you to approach your lover. {\em Why don't you think positively about it without thinking too pessimistically?}'' (3) ``Why don't you tell your lover that you usually do not say what you are thinking. $\cdots$ {\em I think that it is important to communicate the feelings to your lover; how you like or care about him/her especially when you are quarreling with each other.}''

The generation of new answers is important for non-factoid answer systems, since they must cope with slight differences in question contexts from those in the corpus.

\paragraph{Online evaluation in ``Love Advice'' service}
Our ensemble network is currently being used in the love advice
service of Oshiete goo \cite{gtc}. The
service uses only the ensemble network to ensure that the service offers
high-quality output free from grammar errors. We input the sequences in
our evaluation corpus instead of the decoder output sequences into the
ensemble network. Our ensemble network then learned the optimum
combination of answer sequences as well as the closeness of the question
and those sequences. As a result, it can construct an answer that
corresponds to the situation underlying the question. In particular,
5,702 answers created by the AI, whose name is Oshi-el (Oshi-el means
teaching angel), using {\em our ensemble network} in reply to 33,062
questions entered from September 6th, 2016 to November 17th, 2019, were judged
by users of the service as {\em good answers}. Oshi-el output good
answers at about twice the rate of the average human responder in
Oshiete-goo who answered more than 100 questions in the love advice
category. Thus, we think this is a good result.

Furthermore, to evaluate the effectiveness of the supplemental information,
we prepared 100 answers that only contained conclusion sentences during
the same period of time. As a result, users rated the answers that
contained both conclusion and supplement sentences as good 1.6 times
more often than those that contained only conclusion sentences.  This
shows that our method successfully incorporated supplemental information
in answering non-factoid questions.

\section{Conclusion}
\label{sec:conclusion} We tackled the problem of conclusion-supplement
answer generation for non-factoid questions, an important task in
NLP. We presented an architecture, ensemble network, that uses an
attention mechanism to reflect {\em the context} of the conclusion
decoder's output sequence on the supplement decoder's output
sequence. The ensemble network also assesses {\em the closeness} of the
encoder input sequence to the output of each decoder and the combined
output sequences of both decoders. Evaluations showed that our
architecture was consistently superior to conventional encoder-decoders
in this task.  The ensemble network is now being used in the ``Love
Advice,'' service as mentioned in the Evaluation section.

Furthermore, our method, NAGM, can be generalized to generate much
longer descriptions other than conclusion-supplement answers. For
example, it is being used to generate Tanka, which is a genre of
classical Japanese poetry that consists of five lines of
words\footnote{https://www.tankakenkyu.co.jp/ai/}, in the following
way. The first line is input by a human user to NAGM as a question, and
NAGM generates second line (like a conclusion) and third line (like a
supplement). The third line is again input to NAGM as a question, and
NAGM generates the fourth line (like a conclusion) and fifth line (like
a supplement).

\bibliographystyle{aaai}
\bibliography{aaai20}

\end{document}